\documentclass[final]{cvpr}

\usepackage{times}
\usepackage{epsfig}
\usepackage{graphicx}
\usepackage{amsmath}
\usepackage{amssymb}
\usepackage{multirow}
\usepackage{caption}
\usepackage[pagebackref=true,breaklinks=true,colorlinks,bookmarks=false]{hyperref}

\def\cvprPaperID{21} 
\def\confYear{CVPR 2021}

\begin{document}

\title{Unsupervised detection of mouse behavioural anomalies\\ using two-stream convolutional autoencoders}

\author{Ezechukwu I Nwokedi\\
School of Computer Science\\
University of Lincoln, UK\\
{\tt\small enwokedi@lincoln.ac.uk}
\and
Rasneer S Bains\\
MRC Harwell Institute\\
Harwell, UK\\
{\tt\small r.bains@har.mrc.ac.uk}

\and
Luc Bidaut\\
School of Computer Science\\
University of Lincoln, UK\\
{\tt\small lbidaut@lincoln.ac.uk}

\and
Sara Wells\\
MRC Harwell Institute\\
Harwell, UK\\
{\tt\small s.wells@har.mrc.ac.uk}

\and
Xujiong Ye\\
School of Computer Science\\
University of Lincoln, UK\\
{\tt\small xye@lincoln.ac.uk}

\and
James M Brown\\
School of Computer Science\\
University of Lincoln, UK\\
{\tt\small jamesbrown@lincoln.ac.uk}

}

\maketitle

\begin{abstract}
   This paper explores the application of unsupervised learning to detecting anomalies in mouse video data. The two models presented in this paper are a dual stream, 3D convolutional autoencoder (with residual connections) and a dual stream, 2D convolutional autoencoder. The publicly available dataset used here contains twelve videos of a single home-caged mice alongside frame level annotations. Under the pretext that the autoencoder only sees normal events, the video data was handcrafted to treat each behaviour as a pseudo-anomaly thereby eliminating them from the others during training. The results are presented for one conspicuous behaviour (\textbf{hang}) and one inconspicuous behaviour (\textbf{groom}). The performance of these models is compared to a single stream autoencoder and a supervised learning model, which are both based on the custom CAE encoder. Both models are also tested on the CUHK Avenue dataset were found to perform as well as some state-of-the-art architectures. 
\end{abstract}

\section{Introduction}

Over the years, the application of machine learning to studies of animal behaviour has gained momentum with increased availability of data and more innovative and robust techniques. Aside from methods which could aid various processes and automate workflows \cite{valletta2017applications}, machine learning models could promote safer and less intrusive means of carrying out research involving animals. As such, many organisations continue to push for their implementation in various animal-centric laboratories and institutes. 

One of the main ethical challenges which is vital to studies involving animals is the identification of welfare concerns. These could take on several forms, but can be broadly classed as either physiological or behavioural anomalies. In behavioural studies, these anomalies refer to any deviations from the perceived normal behaviour by the subject. For instance, review studies by \cite{lezak2017behavioral} show that rodents can exhibit footshock, flinch, hyper/hypoactivity and forced swim behaviours (amongst a host of others) as symptoms of varying anxiety disorders. However, the definition of what is classed as a behavioural anomaly widely varies across research.

For classification of specific activities from videos, several supervised algorithms have been proposed over the years \cite{karpathy2014large, simonyan2014two, tran2015learning}. However, such models require fully labelled data which, for videos, is an arduous and labour-intensive task. Furthermore, this may not be a viable option in scenarios where events are not clearly defined. However, autoencoders and generative models have been found to be very effective for semi-supervised and unsupervised anomaly detection. Based on the construct in \cite{srivastava2015unsupervised}, the authors of \cite{zhao2017spatio} built a 3D convolutional autoencoder which uses two decoders (one for reconstructing input and the other for predicting future sequences) and achieved high accuracy detection. In addition, \cite{fu2018novel} also achieved good anomaly detection results using a 3D convolutional autoencoder for spatial encoding-decoding, and a 2D convolutional long short-term memory for the temporal encoding-decoding. Furthermore, \cite{ravanbakhsh2017abnormal} proved the effectiveness of conditional generative adversarial networks (GAN) in spotting anomalies by learning only normal spatio-temporal representations.  

\begin{figure}[h]
\centering
\includegraphics[width=0.3\textwidth]{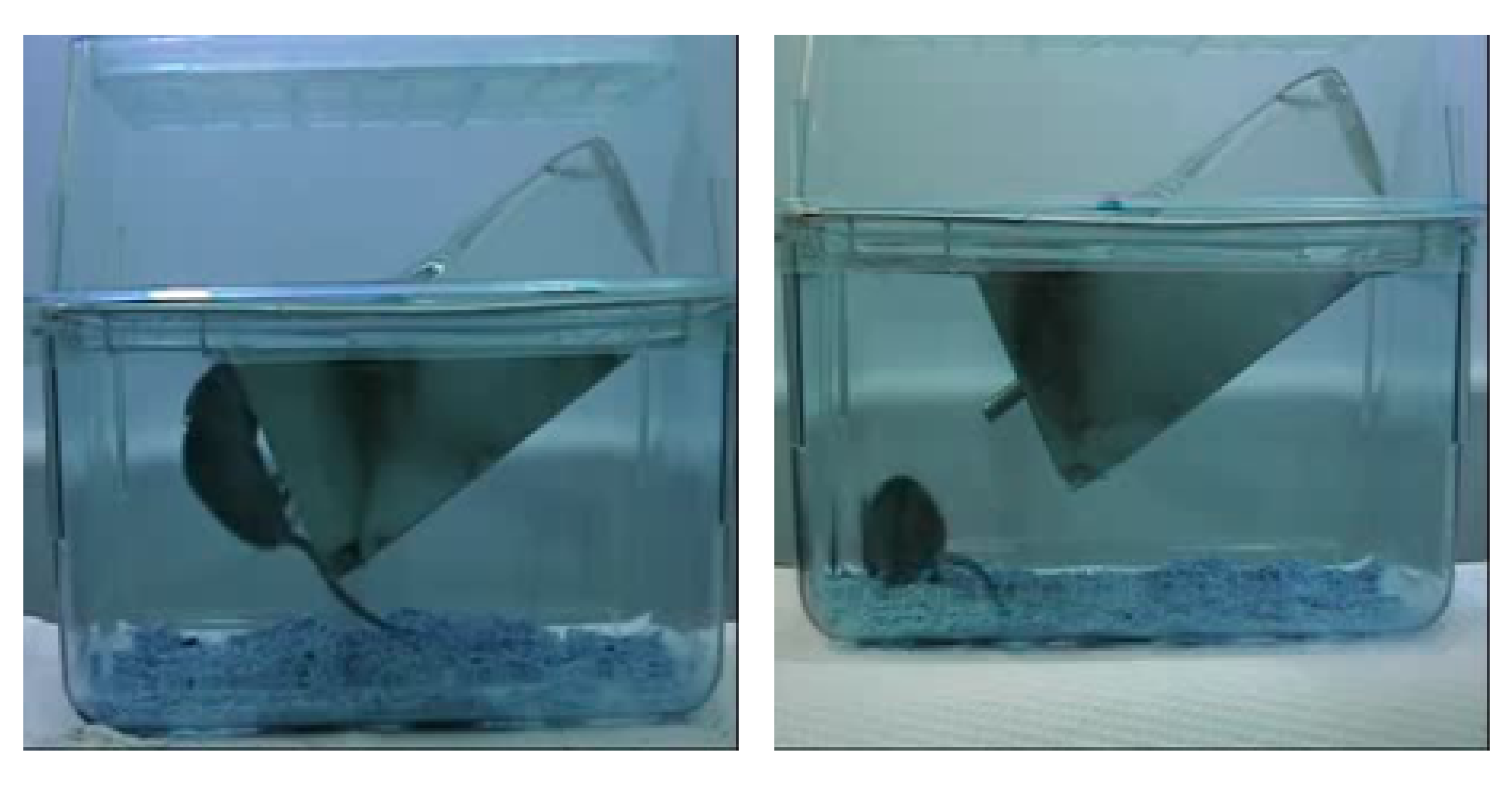}
\caption{Example frames of the \textit{hang} (left) and \textit{groom} (right) pseudo-anomalous behaviours. Dataset from \cite{jhuang2010automated}.}
\label{fig:hang_groom}
\end{figure}

The focus of this paper is to present a proof-of-concept for anomaly detection in mouse videos using autoencoders. This is achieved using two autoencoders to detect behavioural anomalies. Unlike common two-stream models, the autoencoders proposed here are fused at the bottleneck (to allow shared representation learning at lower dimensions) and then separated again, like in \cite{wu2019deep}. The dataset used contains twelve videos of over 10 hours in length, of a individually housed mice with annotations for eight different behaviours \cite{jhuang2010automated}. The annotated behaviours are \textit{drink, eat, groom, hang, rear, rest, micromovement} and \textit{walk}. In our case, the detection models are applied to detect one conspicuous behaviour (\textit{hang}) and one less conspicuous behaviour (\textit{groom}) as pseudo-anomalies (Figure \ref{fig:hang_groom}). Comparisons are also made between these two stream models, a single stream CAE and a fully supervised 2D convolutional network. Additionally, the performance of the two stream models on a public anomaly dataset is presented. 

\section{Methods}

\subsection{Architectures}
The two models proposed are a 3D residual autoencoder (RAE) and a time-distributed 2D convolutional autoencoder (CAE). The 3D RAE is a deep fusion model made up of 3D convolutions and residual connections within both the encoding and decoding sub-models. This model bears some similarities to the work by \cite{wu2019deep} however theirs does not make use of residual connections. This ensures that residual features are enforced at each stage of the encoding/decoding process, thus preserving the bulk of the information. This is essential because most behaviours being treated as anomalies are best observed in their dynamic representations \cite{nguyen2019applying} and not a point or instance occurrence.  

\begin{figure*}[h!]
\centering
\includegraphics[width=\textwidth]{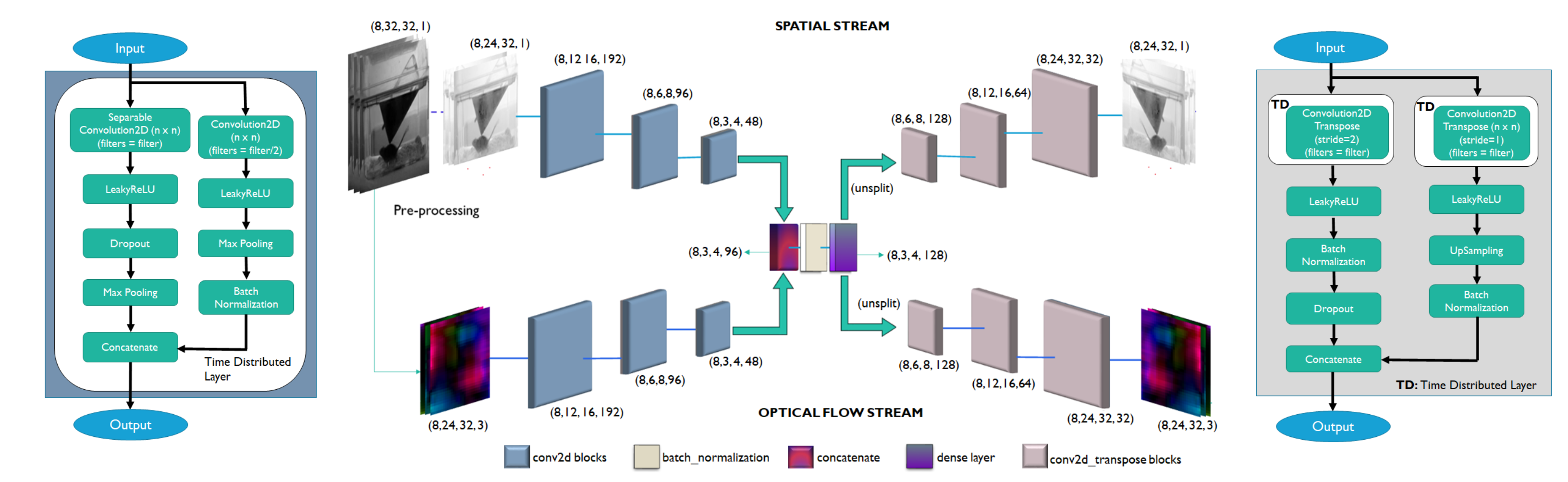}
\caption{\label{fig:cae}The two-stream convolutional autoencoder (CAE) architecture.}
\end{figure*}

On the other hand, the dual CAE was made to thoroughly map spatial features of each of the video frames. By simultaneously applying 2D convolutions to several video frames at a time, the network operates on spatio-temporal cubes like a 3D model. Both models have dual streams that are fused midway. This network was created to push the limits of spatial encoder-decoder mapping and compare the performance against its 3D counterpart (Figure \ref{fig:cae}). The single stream CAE and supervised model, used for comparison, make use of only the grayscale images. 

\subsection{Data \& pre-processing}
The training data for the two fusion models is comprised of the raw video and optical flow images. The video frames are initially resized 32x32 grayscale images, and brightened via gamma correction to deal with variations in illumination and hue.  This allows for a cleaner dense optical flow \cite{farneback2003two} to be computed between successive frames. Both grayscale and dense optical flow images are augmented using horizontal flips and then normalised between 0 and 1. The images are also centrally cropped to 24x32 to remove some redundant regions. Finally, with the aid of the annotations, frames related to the chosen anomalous class are removed from the training data and reshaped into a 4D tensors having 8 frames per sequence.  

\subsection{Training and testing}
Despite having twelve videos available in the dataset, only six are used in training and validating the models to help reduce the heavy imbalance which exists between the labels (detailed distribution of video frames to their classes available in \cite{nguyen2019applying}). Continuous strides of five frames (or approximately one-fifth of every second) and no overlaps were also applied throughout the training data to further reduce its size. For the test, half of one of the videos (whose duration is $>$ 1 hour) was used, with no skips included. This was chosen because it fulfilled the requirement of having all the annotated behaviours. The final training set for the \textit{hang} behaviour had 9577 sequences while the validation set had 2395 sequences. The final data for the \textit{groom} behaviour had 7312 and 1828 sequences for training and validation, respectively. The testing data had 6989 sequences. 

\subsection{Hyperparameters and loss function}

\begin{table}[h!]
\begin{tabular}{ll|l|l|}
\cline{3-4}
                                                                                                       &              & \textit{\begin{tabular}[c]{@{}l@{}}RAE\end{tabular}} & \textit{\begin{tabular}[c]{@{}l@{}}CAE\end{tabular}} \\ \hline
\multicolumn{2}{|l|}{\textit{Loss function}}                                                                          & MSE                                                                      & MSE                                                                                        \\ \hline
\multicolumn{1}{|l|}{\multirow{2}{*}{\textit{\begin{tabular}[c]{@{}l@{}}Loss\\ weights\end{tabular}}}} & \textit{Image}        & 0.75                                                                     & 1.0                                                                                        \\ \cline{2-4} 
\multicolumn{1}{|l|}{}                                                                                 & \textit{Optical flow} & 1.0                                                                      & 1.0                                                                                        \\ \hline
\multicolumn{2}{|l|}{\textit{Optimiser}}                                                                              & Adam                                                                     & Adam                                                                                       \\ \hline
\multicolumn{2}{|l|}{\textit{Learning rate (lr)}}                                                                     & 0.0002                                                                   & 0.0002                                                                                     \\ \hline
\multicolumn{2}{|l|}{\textit{Momentum}}                                                                               & 0.5                                                                      & 0.5                                                                                        \\ \hline
\multicolumn{2}{|l|}{\textit{Epochs}}                                                                                 & 120                                                                      & 60                                                                                         \\ \hline
\multicolumn{2}{|l|}{\textit{Batch size}}                                                                             & 64                                                                       & 32                                                                                         \\ \hline
\multicolumn{2}{|l|}{\textit{\begin{tabular}[c]{@{}l@{}}Dilation rate\\ (convolutions)\end{tabular}}}                 & 1                                                                        & \begin{tabular}[c]{@{}l@{}}2\\ (last two layers)\end{tabular}                              \\ \hline
\end{tabular}
\caption{\label{table:hyper}Hyperparameters used for model training.}
\end{table}

The hyperparameters used in both models can be found in Table \ref{table:hyper}. We utilise a standard mean squared error (MSE) loss function for training:

\begin{equation}
    \mathcal{L}(x, \hat{x}) = \frac{1}{n} \sum{} (x - \hat{x})^2
\end{equation}
 
where is the input $x$ is either the raw video/optical flow and $\hat{x}$ is the decoded reconstruction. The loss $\mathcal{L}$ is computed for each stream independently and a weighted sum computed to produce a combined loss $\mathcal{L}_{d}$. Unlike the CAE, the RAE was found to perform best with a loss-weight ratio of 0.75:1.0 for the grayscale and dense flow images, respectively. In addition, homogenous kernels of (3x3x3) were used throughout the 3D RAE residual layers while the CAE used non-homogenous kernels (3x3, 5x5, etc.) in its 2D convolutions.  

\subsection{Metrics}
Regularity scores were computed via the $\ell^2$ norm of the MSE between the input data and their reconstructions, averaged across both streams. The area under the receiver operator characteristic curve (AUC) is one of the most widely used metrics used in evaluating anomaly detection and hence, was adopted in this paper for all the models. 

\subsection{Equipment}
All code was written in Python 3.8 using the TensorFlow (Keras) library \cite{chollet2015keras}. All models were trained on a single NVIDIA TITAN V GPU. 

\section{Results}

\subsection{Performance on benchmark dataset}
The benchmark used used to test the models was the CUHK Avenue dataset \cite{lu2013abnormal}. Besides normalisation, no additional pre-processing was carried out on this dataset. The same hyperparameters used on the mouse dataset were also used here. The CAE architecture achieved better AUCs than the 3D RAE overall. The AUCs achieved (Table \ref{table:bench}) are in fact on par with those achieved by mainstream anomaly detection models, even better in some cases. This lends credence to the validity of the proposed models and the ideology behind them for detecting anomalies. 
\begin{table}[]
\centering
\begin{tabular}{l|l}
\textit{Method}                      & \textit{AUC} \\ \hline
2D CAE                               & 0.702        \\
Detection at 150 FPS                 & 0.809        \\
Two-stream 3D Conv AE                & 0.866        \\
STAE (spatial only)                  & 0.771        \\
STAE (optical flow only)             & 0.809        \\
DSTN                                 & 0.879        \\ \hline
Our single-stream CAE (spatial only) & 0.784        \\
Our two-stream 3D RAE                & 0.826        \\
Our two-stream CAE                   & 0.832       
\end{tabular}

\caption{\label{table:bench}Comparison of methods on the CUHK Avenue dataset.}

\end{table}

\subsection{Performance on less conspicuous behaviour}
The \textit{groom} behaviour is, unlike \textit{hang}, a less perceptible mouse behaviour. From a static point of view, it can easily be confused for a number of others, including micromovement and eating. The challenge in identifying this behaviour was also observed by the authors of the dataset [3], who recorded 57\% accuracy in human detection and about 70\% accuracy from their supervised approach. Thus, the results achieved were not desirable, but still marginally better than a random guesswork (AUC=0.5). 

\begin{figure}
\centering
\includegraphics[width=0.5\textwidth]{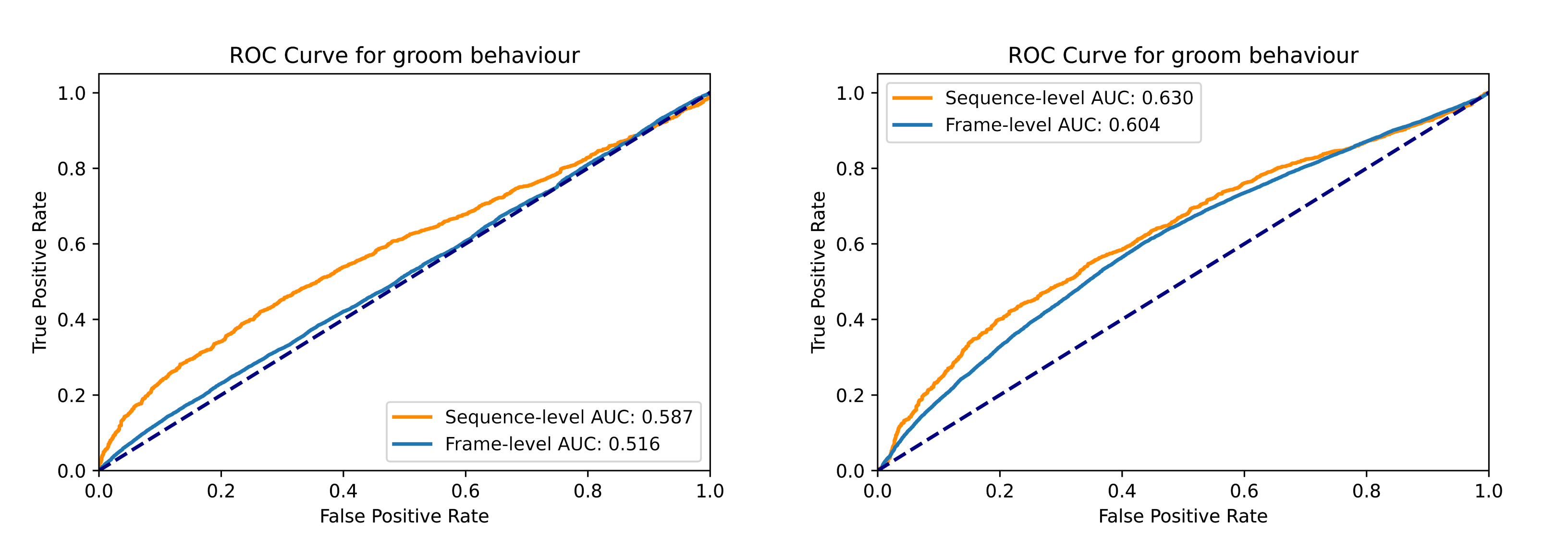}
\caption{Two-stream models on \textit{groom}: residual autoencoder (left) and convolutional autoencoder (right)}
\label{fig:Dual_Stream_RAE_and_CAE_on_groom.png}
\end{figure}

\subsection{Performance on more conspicuous behaviour}
Here, the two-stream models achieved excellent results having AUCs of over 0.9. Though anomalies are not always as noticeable as \textit{hang}, the performance of the models demonstrates that the two-stream autoencoders can detect obvious behavioural mouse anomalies with high peformance. 

\begin{figure}
\centering
\includegraphics[width=0.5\textwidth]{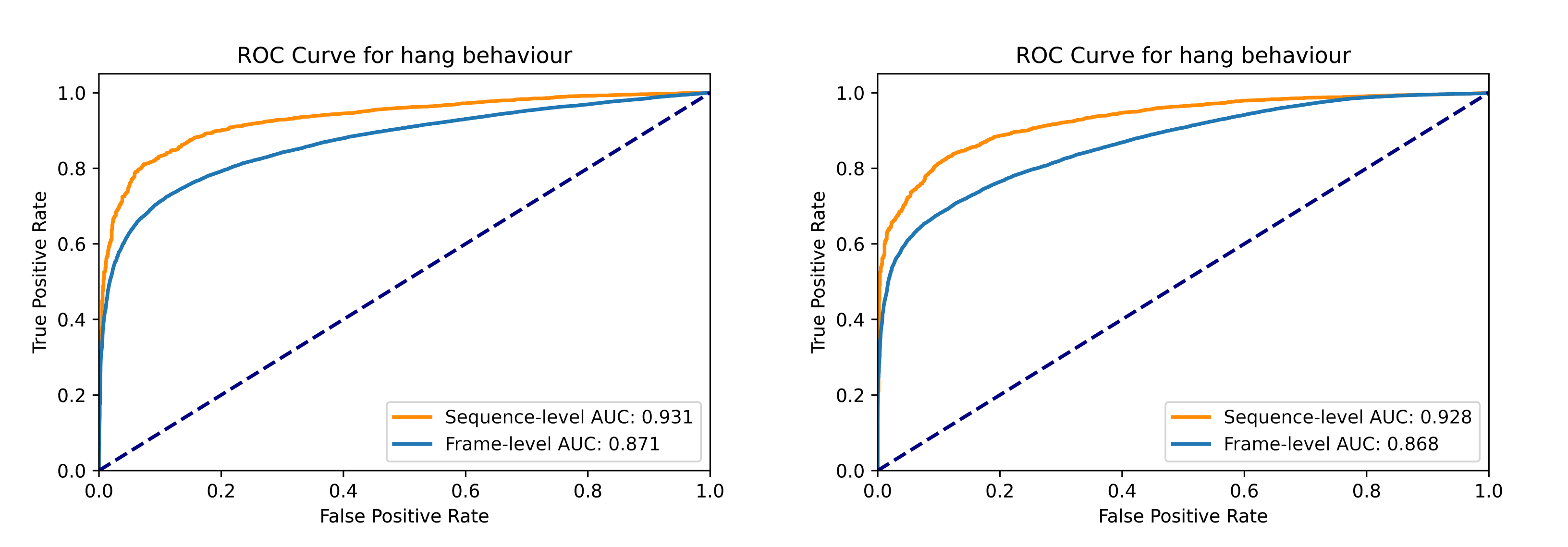}
\caption{Two-stream models on \textit{hang}: residual autoencoder (left) and convolutional autoencoder (right)}
\label{fig:Dual_Stream_RAE_and_CAE_on_groom.png}
\end{figure}

\subsection{Performance of alternate networks}
Two alternative models were also developed, derived from the dual CAE, to compare their performance with the two stream models; these were done for supervised and unsupervised learning modes. The unsupervised model explores how well the single stream autoencoder works using spatial data (grayscale). It made use of the same hyperparameters as the two-stream CAE. The supervised model made use of the CAEs encoder and an appended fully connected network. This was trained for 120 epochs using stochastic gradient descent and a learning rate of 0.0005.  

The supervised model achieved AUCs of 0.99 and 0.66 on \textit{hang} and \textit{groom} respectively, while the one-stream CAE had 0.959 and 0.644. In comparison to the dual stream models, these two approaches achieved higher AUCs on both behaviours. This was expected of the supervised model because it has the benefit of the annotations to aid mapping. Furthermore, its effectiveness also confirms the ability of the proposed 2D CAE’s encoder in spatial representation. However, the improved performance of a one-stream model was unexpected, and suggests that further pre-processing may be needed to the video prior to calculation of optical flow, or refinement of the dual loss weighting.

\section{Conclusion}
The two-stream models treated in this paper prove the applicability of fully unsupervised autoencoders in detecting mouse behaviours. In the initial validation study, the models achieved results comparable with the state-of-the-art on the CUHK Avenue benchmark dataset. While the problem formulation for mouse anomalies was artificial, it demonstrated the utility of the proposed approach for genuine anomalies. This study provides a foundation for more intensive investigation into improvements from both the data processing and the model tunning ends. For the next steps, exploring unsupervised forms of even deeper networks and other generative models are promising alternatives to detecting subtle behavioural anomalies. 
\section{Acknowledgements}
This research was funded by the National Centre for the Replacement, Refinement \& Reduction of Animals in Research (NC/T002050/1).

{\small
\bibliographystyle{ieee_fullname}
\bibliography{egbib}
}

\end{document}